\documentclass[a4paper,UKenglish,cleveref, autoref, thm-restate]{lipics-v2021}
\usepackage{multirow}
\usepackage{gensymb}                        % enable some symbols

\title{Evaluating the Effectiveness of Large Language Models in Representing Textual Descriptions of Geometry and Spatial Relations} 

\titlerunning{Evaluating the Effectiveness of LLMs in Spatial Representations} %TODO optional, please use if title is longer than one line

\author{Yuhan Ji}{GeoDS Lab, Department of Geography, University of Wisconsin-Madison, USA}{yuhan.ji@wisc.edu}{https://orcid.org/0000-0002-4426-6105}{}%TODO mandatory

\author{Song Gao}{GeoDS Lab,  Department of Geography, University of Wisconsin-Madison, USA}{song.gao@wisc.edu}{https://orcid.org/0000-0003-4359-6302}{}

\authorrunning{Ji and Gao} %TODO mandatory. First: Use abbreviated first/middle names. Second (only in severe cases): Use first author plus 'et al.'

\Copyright{Yuhan Ji and Song Gao} %TODO mandatory, please use full first names. LIPIcs license is "CC-BY";  http://creativecommons.org/licenses/by/3.0/

\ccsdesc[500]{Computing methodologies~Artificial intelligence}

\keywords{LLMs, foundation models, GeoAI} %TODO mandatory; please add comma-separated list of keywords

\relatedversion{} %optional, e.g. full version hosted on arXiv, HAL, or other respository/website
\supplement{}%optional, e.g. related research data, source code, ... hosted on a repository like zenodo, figshare, GitHub, ...

\acknowledgements{The authors would like to acknowledge the support from the H.I. Romnes Fellowship, National Science Foundation (No. 2112606) and Arity.}%optional

\nolinenumbers %uncomment to disable line numbering

\hideLIPIcs  %uncomment to remove references to LIPIcs series (logo, DOI, ...), e.g. when preparing a pre-final version to be uploaded to arXiv or another public repository

\EventEditors{Roger Beecham, Jed A. Long, Dianna Smith, Qunshan Zhao, and Sarah Wise}
\EventNoEds{5}
\EventLongTitle{12th International Conference on Geographic Information Science (GIScience 2023)}
\EventShortTitle{GIScience 2023}
\EventAcronym{GIScience}
\EventYear{2023}
\EventDate{September 12--15, 2023}
\EventLocation{Leeds, UK}
\EventLogo{}
\SeriesVolume{277}
\ArticleNo{72}
\begin{document}

\maketitle

\begin{abstract}
\footnote{a preprint and the final version will be available in the Proceedings of the 12th International Conference on Geographic Information Science (GIScience 2023) \url{https://www.giscience.org/}.}This research focuses on assessing the ability of large language models (LLMs) in representing geometries and their spatial relations. We utilize LLMs including GPT-2 and BERT to encode the well-known text (WKT) format of geometries and then feed their embeddings into classifiers and regressors to evaluate the effectiveness of the LLMs-generated embeddings for geometric attributes. The experiments demonstrate that while the LLMs-generated embeddings can preserve geometry types and capture some spatial relations (up to 73\% accuracy), challenges remain in estimating numeric values and retrieving spatially related objects. This research highlights the need for improvement in terms of capturing the nuances and complexities of the underlying geospatial data and integrating domain knowledge to support various GeoAI applications using foundation models. 

\end{abstract}

\section{Introduction}
\label{sec:intro}
    
Deep learning methods have exhibited great performance to tackle many challenging tasks in geographical sciences~\cite{reichstein2019deep,janowicz2020geoai}. However, the models often depend on handcrafted features for specific downstream tasks, thus being hard to be generalized into different tasks. The emergence of representation learning largely mitigated the issue by decomposing the learning process into two steps (task-agnostic data representation and downstream task)~\cite{bengio2013representation}. 
%First, learn the task-agnostic representation that captures the underlying structure of the input data, and second, input the learned representation (i.e., embedding) to perform a downstream task~\cite{bengio2013representation}. 
Therefore, an effective location-based representation should preserve key spatial information (e.g., distance, direction, and spatial relations) and make classifiers or other predictors easy to extract useful  knowledge~\cite{mai2022review}. In geospatial artificial intelligence (GeoAI) research, although the geospatial data are usually well-formatted and can be readily understood by GIS software, not all of them can be directly integrated into a deep learning model. 
    
The success of ChatGPT has been a milestone that attracts the general public's attention to Large Language Models (LLMs). With tons of parameters trained on a large text corpus, LLMs have learned profound knowledge across many domains. Other well-known LLMs include the Bidirectional Encoder Representations from Transformers (BERT)~\cite{devlin2018bert}, the Generative Pre-trained Transformer (GPT) series \cite{radford2019language, brown2020language}, etc.
%Their variants and alternative models. They differ in the network architectures, size, training data, and intended use, which may yield a discrepancy in the learned linguistic patterns and semantics. 
Despite the differences in network architectures, these LLMs can achieve state-of-the-art performance on natural language processing (NLP) benchmarks. 
Consequently, researchers have begun the early exploration of integrating LLMs into GIS research, such as geospatial semantic tasks~\cite{mai2023opportunities} and automating spatial analysis workflows~\cite{li2023autonomous}. These studies have demonstrated the ability of LLMs to understand and reason about geospatial phenomena from a semantic perspective as learned from human discourse or formalized programming instructions. In contrast, accurate geometries and spatial relations in GIS are not necessarily expressed in natural languages. Therefore, it can be challenging for LLMs to reconstruct the physical world solely from the textual description of these building blocks, which is the motivation of this research. 

In GIScience, spatial relations refer to the connection between spatial objects regarding their geometric properties~\cite{guo1998spatial} 
% while topological relations are defined as spatial relations that are preserved under transformations such as rotation and scaling~\cite{clementini1996model}
, which play an important role in spatial query, reasoning and question-answering. Using natural language to describe spatial relations is essential for humans to perceive our surroundings and navigate through space. Attempts have been made to formalize the conversion between quantitative models and qualitative human discourse \cite{cohn2001qualitative}. For topological spatial relations, the RCC-8 (region connection calculus \cite{randell1992spatial}) and the Dimensionally Extended 9-intersection (DE-9IM) model~\cite{egenhofer2005reasoning} are widely used. %RCC-8 is a set of eight jointly exhaustive and pairwise disjoint relations defined for regions. The basic relations include predicates like (externally) connected, disconnected, (partially) overlaps, tangential, and nontangential, which turn out to be cognitively adequate to be well distinguished by human \cite{renz1998spatial}.  Dimensionally Extended 9-intersection (DE-9IM) model \cite{egenhofer2005reasoning} defines a $3\times3$ matrix that describes the binary relation between the exterior, boundary, and interior of two objects, thus being able to distinguish at most 512 relations though some of them are impossible. 
Based on the DE-9IM model, five predicates are named by \cite{clementini1996model} for complex geometries, including \textit{crosses, disjoint, touches, overlaps, within}. On top of them, the Open Geospatial Consortium (OGC) further added the predicates \textit{equals, contains, intersects} for computation convenience. In addition, predicates can also be used to describe the distance or direction between a subject and an object. Fuzzy logic can also be adopted to convert precise metrics into narrative predicates such as \textit{near} and \textit{far}~\cite{wang2023spatial}. %Physical properties and bounding boxes of the objects can jointly be used to the threshold for the conversion \cite{hu2016natural, abella1993qualitatively}.

% difficulty - vagueness & context-dependent
However, there remains a gap between the contextual semantics of predicates in everyday language and the abovementioned formalization procedures, yielding disagreement and vagueness in the understanding. It is yet to be determined whether the LLMs can fully capture how people describe spatial objects with predicates in natural language. If so, how we can leverage such knowledge to represent geospatial contexts with LLMs.

\section{Methodology}
\subsection{Workflow}
% dataset -> model -> embeddings -> geometry task 
% concatenate
% spatial relation task
% classification/regression network architecture

\begin{figure}[h]
    \centering
    \includegraphics[width=.8\linewidth]{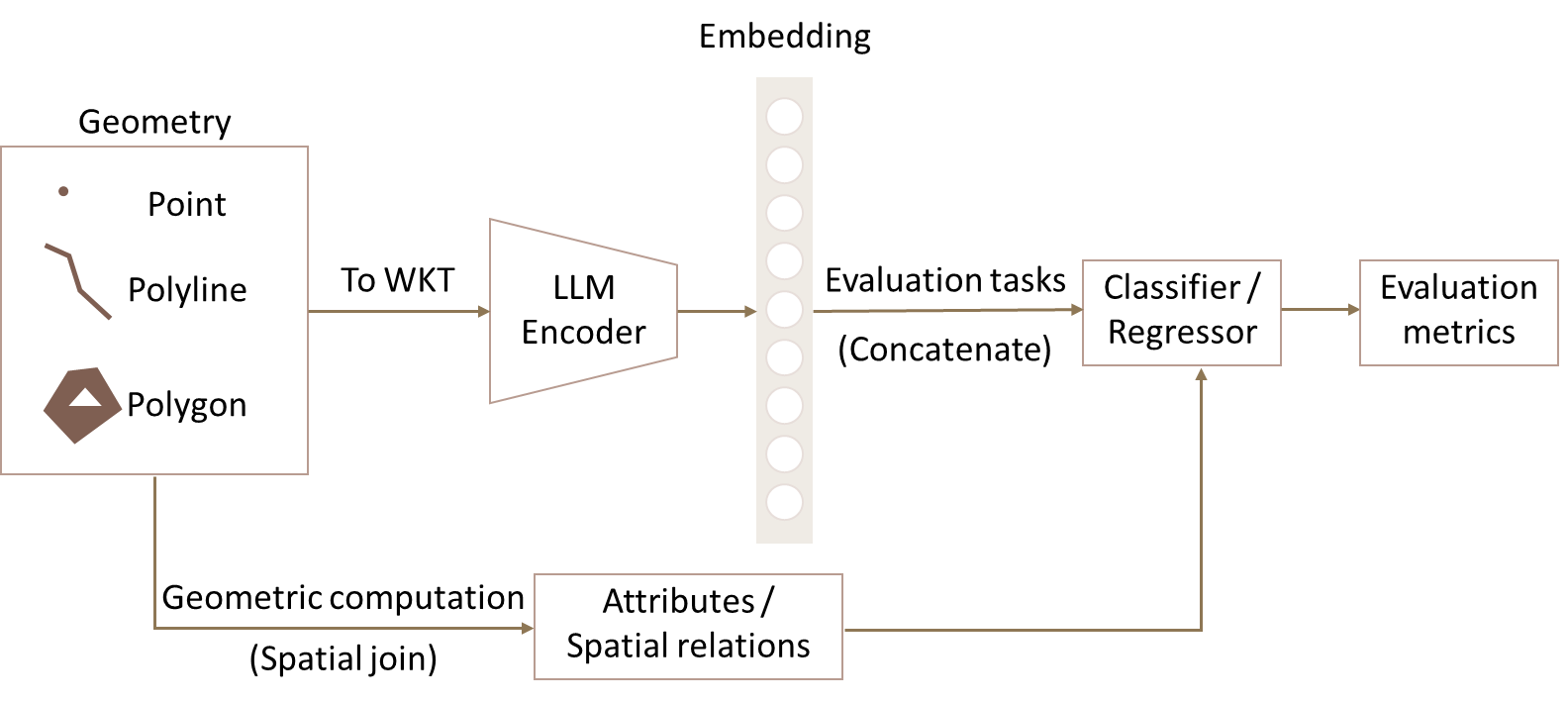}
    \caption{The evaluation workflow of this research}
    \label{fig:framework}
\end{figure}

This research focuses on assessing the ability of LLMs in representing geometries and their spatial relations through a set of downstream tasks. Figure \ref{fig:framework} illustrates the workflow we employed, which consists of three primary modules. The first module utilizes a GIS tool to extract the attributes, such as geometry type, centroid, and area, of individual geometries and their spatial relations, including predicates and distances between pairs of geometries. The second module applies LLMs to encode the well-known text (WKT) format of geometries, e.g., LINESTRING (30 10, 10 30, 40 40), which includes the geometry type and the ordered coordinates whereas the map projection is not considered in this work. 
Finally, the obtained embeddings from LLMs, along with the ground-truth attributes or spatial relations, are fed into classifiers or regressors to evaluate the effectiveness of the LLMs-based embeddings. 

\subsection{Notation}

The notations used in this paper are listed in Table \ref{tab:notation}.

\begin{table}[ht]
\footnotesize
    \centering
    \vspace{-1em}
    \caption{Notations}
    \begin{tabular}{p{.2\linewidth}|p{.75\linewidth}}
    \hline
        Notation &  Description\\
        \hline
        $g$ & A geometry instance (e.g. Point, LineString, and Polygon) that can be processed in GIS tools \\
        $WKT(g)$ & The WKT format of $g$ \\
        $Enc(g)$ & The location encoding of $g$ using a LLM model to encode $WKT(g)$\\
        $Type(g)$ & The geometry type of $g$ \\
        $Centroid(g)$ & The centroid of $g$ \\
        $Area(g)$ & The area of $g$ \\
        $rel$ & A predicate that can be used to represent the spatial relation, which is one of \{equals, disjoint, intersects, crosses, touches, contains, within, overlaps\}, as defined by OGC and implemented in GeoPandas.\\ 
        $Rel(g_i, g_j)$ & The spatial relation between the subject $g_i$ and the object $g_j$ \\
        $Dist(g_i, g_j)$ & The minimum euclidean distance between two objects $g_i$ and $g_j$ \\
        $\left[Enc(g_i);Enc(g_j)\right]$ & The concatenation of the embeddings of $g_i$ and $g_j$ \\ 
        $Enc(rel, g)$ & The embedding of the short phrase $rel+ WKT(g)$. For example, ``within Polygon ((0 0, 0 1, 1 1, 1 0, 0 0))'' \\
    \hline
    \end{tabular}
    \label{tab:notation}
\end{table}

\subsection{Evaluation Tasks}
The downstream tasks are designed for deriving the geometric attributes or identifying spatial relations, as described in Table \ref{tab:tasks}. The targets of Tasks 1-5 are straightforward, that is, to train a neural network classification/regression model that can best approximate the ground-truth values computed from a GIS tool. All of these tasks use a Multilayer Perceptron (MLP) as the classifier or regressor.
Task 6 aims to investigate whether a geometry $g_i$ can be predicted based on its neighbor $g_j$ and their spatial relation $Rel(g_i, g_j)$. 
% During training, the model is trained to minimize the similarity between the actual embedding $Enc(g_i)$ and the derived embedding $Enc(rel, g_j)$. 
We employ the nearest neighbor retrieval approach to evaluate whether LLMs have learned the meaning of spatial predicates properly. During inference, given an object $g_j$ and a spatial relation $rel$, we retrieve the top-k nearest neighbors of $Enc(rel, g_j)$ and examined whether they belong to the set of subjects $\{g_i| Rel(g_i, g_j)=rel\}$. This approach assesses the ability of the LLMs to relate geographic objects through spatial predicates.

\begin{table}[ht]
%\scriptsize
\footnotesize
\centering
    \caption{Evaluation Tasks}
    \begin{tabular}{|>{\centering\arraybackslash}p{.1\linewidth}|
    >{\arraybackslash}p{.25\linewidth}|
    >{\arraybackslash}p{.12\linewidth}|
    >{\centering\arraybackslash}p{.18\linewidth}|
    >{\centering\arraybackslash}p{.22\linewidth}|}
    \hline
        Task & Subtask  & Model type & Input & Target\\
        \hline
        \multirow{3}{\linewidth}{Geometric attributes} & T1: Geometry type & Classification & $Enc(g)$ & $Type(g)$\\
        % \cline{2-5}
        & T2: Area computation& Regression & $Enc(g)$ & $Area(g)$\\
        % \cline{2-5}
        & T3: Centroid derivation& Regression & $Enc(g)$ & $Centroid(g)$\\
        \hline
        \multirow{3}{\linewidth}{Spatial relations} & T4: Spatial predicate & Classification & $\left[Enc(g_i);En(g_j)\right]$ & $Rel(g_i, g_j)$ \\
        % \cline{2-5}
        & T5: Distance measure & Regression &$\left[Enc(g_i);En(g_j)\right]$ & $Dist(g_i, g_j)$\\
        % \cline{2-5}
        & T6: Location prediction & Retrieval & $Enc(rel, g_j)$ & $\{g_i|Rel(g_i, g_j)=rel\}$\\ 
        \hline
    \end{tabular}
    \label{tab:tasks}
\end{table}

\section{Experiments}
\subsection{Dataset and Preprocessing}
Since there is no available benchmark dataset, we constructed real-world multi-sourced geospatial datasets for our case study in Madison, Wisconsin, United States. We downloaded the OpenStreetMap road network data (including links and intersections) using \textit{OSMnx} \footnote{http://osmnx.readthedocs.io/}, points of interest (POIs) categorized by \textit{SLIPO} \footnote{http://slipo.eu/}, and \textit{Microsoft Building Footprints} \footnote{http://www.microsoft.com/maps/building-footprints}. Our evaluation tasks focus on the spatial objects with \textit{Point, LineString}, and \textit{Polygon} geometry types and assessing their spatial relations, respectively. The datasets are created as follows. 

1) For each geometry type, we randomly select 4,000 samples, including 2,000 road intersections and 2,000 POIs for \textit{Point} data, 4,000 road links for \textit{LineString} data, and 4,000 building footprints for \textit{Polygon} data. In total 12,000 samples are used for performing the downstream tasks. The area and centroid of each polygon are also computed.

2) For the spatial predicate \textit{disjoint}, we randomly generate pairs of geometries and check whether their spatial relation is disjoint. For other predicates, we identify spatially related objects using spatial join. Given each combination of subject/object geometry type and their spatial predicate, we keep 400 triplets (subject, predicate, object) for each category for the task of predicate prediction and distance measure. Then we compute the minimum distance between the subjects and the objects.  

% 3) We further sample 20\% of the data from step 1) to be used as referenced objects in the task of location recognition. We compute the neighbors of each geometry in the entire dataset using a buffer radius of 0.003\textdegree, which will be used as subjects in the task of spatial relation prediction. The predicate of ``disjoint'' is replaced by ``disjoint but near''.

% 4) We compute the spatial relation between subjects and objects using the predicates defined by OGC. For a given object and a predicate, we keep up to 5 subjects (if exists). Then we compute the minimum distance between the subjects and the object. 
3) We further construct data for the task of location prediction. In addition to the subjects and objects that are spatially joined in step 2), we also relate neighboring disjoint geometries using a buffer radius of 0.003\textdegree. The predicate of ``disjoint'' is replaced by ``disjoint but near''. For each predicate except \textit{disjoint}, we select 200 objects of each geometry type that are related to more than 5 subjects by the same predicate.

All the computations are performed by using the \textit{GeoPandas} package in Python. We consider the predicates of \textit{crosses, disjoint (but near), touches, overlaps, within, equals, contains} in this work but not 
\textit{intersects} as it is the opposite of \textit{disjoint}. The data for downstream tasks are further split into 80\% training, 5\% validation, and 15\% test sets.
%In addition, since we focus on the effectiveness of encoding coordinates using LLMs in this work, the datasets are not projected. 

\subsection{Encoding}
In this work, we perform the evaluation tasks based on two LLMs: GPT-2 and BERT. Due to the computational and memory resources required to train and use the models, GPT-2 and BERT have a maximum input sequence length (i.e., 1024 and 512 tokens respectively). Therefore, a sliding window approach is employed to tackle the issue as the WKT of \textit{LineString} and \textit{Polygon} types can exceed the length limitation. The long input sequences are broken down into smaller segments of 512 tokens with an overlap of 256 tokens between adjacent segments. Each segment is processed by the LLMs separately. We then take the average of the token embeddings to generate the final embedding for the whole sequence of geometries.

\subsection{Training MLPs}
As we hypothesize that the learned embeddings from LLMs can be effectively utilized in downstream geometry-related tasks, we use a simple neural network architecture (i.e., MLP) across all tasks. Specifically, the input layer of the MLP is the embedding layer generated from LLMs, followed by a dropout layer for regularization purposes. Following the dropout layer is a single hidden layer, which employs the Rectified Linear Unit (ReLU) activation function. Finally, the MLP is concluded with the output linear layer. The number of neurons in the output layer varies depending on the specific task. 

To facilitate the training process, we apply a logarithmic function to the target values for the area computation and distance measure tasks. In the centroid derivation task, we use the min-max normalization for the target values. The loss function combines the Mean Squared Error (MSE) on both the transformed and original scales. However, for reporting the performance, we only use the original scale of the target values.

\subsection{Results}

% overall evaluation
As shown in Table \ref{tab:results}, the performance of the downstream tasks based on the embeddings generated by GPT-2 and BERT are similar, which can be understood from the similarity in their subword tokenization and transformer-based architecture.

\begin{table}[h]
\footnotesize
\caption{LLMs Performance Comparison} 
\begin{tabular}{|ll|l|ll|ll|}
\hline
\multicolumn{2}{|c|}{\multirow{2}{*}{Tasks}}                                        & \multirow{2}{*}{Metric}             & \multicolumn{2}{c|}{GPT-2}                               & \multicolumn{2}{c|}{BERT}                               \\ \cline{4-7} 
\multicolumn{2}{|l|}{}                                                             &                                     & \multicolumn{1}{c|}{Validation}  & Test                      & \multicolumn{1}{c|}{Validation}  & Test                      \\ \hline
\multicolumn{1}{|l|}{T1: Geometry type}               &                       & Accuracy(\%)                         & \multicolumn{1}{c|}{100}    & 100                       & \multicolumn{1}{c|}{100}    & 100                       \\ \hline
\multicolumn{1}{|l|}{\multirow{2}{*}{T2: Area computation}}    & All geometries           & \multirow{2}{*}{MAPE(\%)}            & \multicolumn{1}{c|}{13124} &    11700                & \multicolumn{1}{c|}{12251} & 10850                    \\ \cline{2-2} \cline{4-7} 
\multicolumn{1}{|l|}{}                                     & Polygon only          &                                    & \multicolumn{1}{c|}{45.1 } &    44.1                & \multicolumn{1}{c|}{40.7} & 41.9                   \\ \hline
\multicolumn{1}{|l|}{T3: Centroid derivation}                  &                       & RMSE                              & \multicolumn{1}{c|}{0.037}  & 0.037                     & \multicolumn{1}{c|}{0.029}  & 0.029                     \\ \hline
\multicolumn{1}{|l|}{\multirow{2}{*}{T4: Spatial predicate}} & Without geometry type & \multirow{2}{*}{Accuracy(\%)}        & \multicolumn{1}{c|}{62.6}   & 65.7                      & \multicolumn{1}{c|}{63.8}   & 68.7                      \\ \cline{2-2} \cline{4-7} 
\multicolumn{1}{|l|}{}                                     & With geometry type    &                                     & \multicolumn{1}{c|}{73.7}   & 71.0                      & \multicolumn{1}{c|}{73.1}   & 72.3                      \\ \hline
\multicolumn{1}{|l|}{T5: Distance measure}                     & Disjoint only         & RMSE                              & \multicolumn{1}{c|}{0.064}  & 0.063                     & \multicolumn{1}{c|}{0.057}    &      0.075                     \\ \hline
\multicolumn{1}{|l|}{T6: Location prediction}                 & \multicolumn{1}{l|}{} & \multicolumn{1}{l|}{Precision@5} & \multicolumn{1}{l|}{N/A}    & \multicolumn{1}{l|}{0.03} & \multicolumn{1}{l|}{N/A}    & \multicolumn{1}{l|}{0.03} \\ \hline
\end{tabular}
\label{tab:results}
\end{table}

For T1-T3, the assessment is conducted on individual geometries. The 100\% accuracy achieved on both the validation and the test dataset of T1 is expected as the geometry type are words that often occur in text documents. Considering the unit of \textit{degree} in longitude and latitude, significant errors (measured by Mean Absolute Percentage Error (MAPE) and Root Mean Square Error (RMSE)) are observed in area and centroid computations, and increasing or reducing the model complexity does not alleviate the issue, suggesting a potential loss of information when averaging the token embeddings or fragmentation of coordinates during tokenization. Training the regressor on all geometries for T2 does not successfully learn that \textit{Point} and \textit{LineString} have an area of 0. Even when training the regressor on \textit{Polygon} separately, the results remain unsatisfactory. In T3, the centroids computed from the high-dimensional embeddings often fall outside the study area. 
T4-T6 evaluates the embeddings' ability to capture spatial relations. One interesting finding is that the spatial predicate can be better predicted when combined with the geometry type, with accuracy increased from 62\%$\sim$68\% to 71\%$\sim$73\%. This can be attributed to the imbalanced spatial relations among different combinations of geometry types. However, the distance measure task T5 still faces challenges in accurately estimating numeric values even when restricted to the ``disjoint'' relation only. The poor performance on T6 shows that even though the LLMs can encode the spatial relations and geometries in a consistent way, generating embeddings using an average approach alone is insufficient to support spatial reasoning and conduct geometric manipulations directly. Therefore, a different design to enhance the function of localizing spatial objects from textual descriptions~\cite{vasardani2013locating} can improve the applications of LLMs in GeoAI.

% Effectiveness in classification tasks
Overall, the results indicate that the LLMs-generated embeddings have encoded the geometry types and coordinates present in the WKT format of geometries. 
However, it should be noted that the performance of the embeddings does not consistently meet expectations across all evaluation tasks. While the LLMs-generated embeddings can preserve geometry types and capture some spatial relations, challenges remain in estimating numeric values and retrieving spatially related objects due to the loss of magnitude during tokenization \cite{frieder2023mathematical}. Despite the possibility of ameliorating the issue by modifying notations or applying chain-of-thought prompting \cite{lample2019deep},
this research highlights the need for improvement in terms of capturing the nuances and complexities of the underlying geospatial data and integrating domain knowledge to support various GeoAI applications using LLMs. 

\bibliography{references}

\begin{thebibliography}{10}

\bibitem{bengio2013representation}
Y.~Bengio, A.~Courville, and P.~Vincent.
\newblock Representation learning: A review and new perspectives.
\newblock {\em IEEE transactions on pattern analysis and machine intelligence},
  35(8):1798--1828, 2013.

\bibitem{brown2020language}
T.~B. Brown et~al.
\newblock Language models are few-shot learners, 2020.

\bibitem{clementini1996model}
E.~Clementini and P.~Di~Felice.
\newblock A model for representing topological relationships between complex
  geometric features in spatial databases.
\newblock {\em Information sciences}, 90(1-4):121--136, 1996.

\bibitem{cohn2001qualitative}
A.~G. Cohn and S.~M. Hazarika.
\newblock Qualitative spatial representation and reasoning: An overview.
\newblock {\em Fundamenta informaticae}, 46(1-2):1--29, 2001.

\bibitem{devlin2018bert}
J.~Devlin, M.-W. Chang, K.~Lee, and K.~Toutanova.
\newblock Bert: Pre-training of deep bidirectional transformers for language
  understanding.
\newblock {\em arXiv preprint arXiv:1810.04805}, 2018.

\bibitem{egenhofer2005reasoning}
M.~J. Egenhofer.
\newblock Reasoning about binary topological relations.
\newblock In {\em Proceedings of the 2nd Symposium on Advances in Spatial
  Databases: SSD'91 Zurich, Switzerland, August 28--30}, pages 141--160.
  Springer, 1991.

\bibitem{frieder2023mathematical}
S.~Frieder, L.~Pinchetti, R.-R. Griffiths, T.~Salvatori, T.~Lukasiewicz, P.~C.
  Petersen, A.~Chevalier, and J.~Berner.
\newblock Mathematical capabilities of chatgpt.
\newblock {\em arXiv preprint arXiv:2301.13867}, 2023.

\bibitem{guo1998spatial}
R.~Guo.
\newblock Spatial objects and spatial relationships.
\newblock {\em Geo-spatial Information Science}, 1(1):38--42, 1998.

\bibitem{janowicz2020geoai}
K.~Janowicz, S.~Gao, G.~McKenzie, Y.~Hu, and B.~Bhaduri.
\newblock Geoai: spatially explicit artificial intelligence techniques for
  geographic knowledge discovery and beyond.
\newblock {\em International Journal of Geographical Information Science},
  34(4):625--636, 2020.

\bibitem{lample2019deep}
G.~Lample and F.~Charton.
\newblock Deep learning for symbolic mathematics.
\newblock {\em arXiv preprint arXiv:1912.01412}, 2019.

\bibitem{li2023autonomous}
Z.~Li and H.~Ning.
\newblock Autonomous gis: the next-generation ai-powered gis.
\newblock {\em arXiv preprint arXiv:2305.06453}, 2023.

\bibitem{mai2023opportunities}
G.~Mai, W.~Huang, J.~Sun, S.~Song, D.~Mishra, N.~Liu, S.~Gao, T.~Liu, G.~Cong,
  Y.~Hu, et~al.
\newblock On the opportunities and challenges of foundation models for
  geospatial artificial intelligence.
\newblock {\em arXiv preprint arXiv:2304.06798}, 2023.

\bibitem{mai2022review}
G.~Mai, K.~Janowicz, Y.~Hu, S.~Gao, B.~Yan, R.~Zhu, L.~Cai, and N.~Lao.
\newblock A review of location encoding for geoai: methods and applications.
\newblock {\em International Journal of Geographical Information Science},
  36(4):639--673, 2022.

\bibitem{radford2019language}
A.~Radford, J.~Wu, R.~Child, D.~Luan, D.~Amodei, I.~Sutskever, et~al.
\newblock Language models are unsupervised multitask learners.
\newblock {\em OpenAI blog}, 1(8):9, 2019.

\bibitem{randell1992spatial}
D.~A. Randell, Z.~Cui, and A.~G. Cohn.
\newblock A spatial logic based on regions and connection.
\newblock {\em KR}, 92:165--176, 1992.

\bibitem{reichstein2019deep}
M.~Reichstein, G.~Camps-Valls, B.~Stevens, M.~Jung, J.~Denzler, and
  N.~Carvalhais.
\newblock Deep learning and process understanding for data-driven earth system
  science.
\newblock {\em Nature}, 566(7743):195--204, 2019.

\bibitem{vasardani2013locating}
M.~Vasardani, S.~Winter, and K.-F. Richter.
\newblock Locating place names from place descriptions.
\newblock {\em International Journal of Geographical Information Science},
  27(12):2509--2532, 2013.

\bibitem{wang2023spatial}
Y.~Wang, H.~Peng, Y.~Xiong, and H.~Song.
\newblock Spatial relationship recognition via heterogeneous representation: A
  review.
\newblock {\em Neurocomputing}, 2023.

\end{thebibliography}

\end{document}